**Enhancing Credit Risk Prediction: A Multi-stage Ensemble Pipeline**

Haibo Wang[1], Jun Huang[2], Lutfu S. Sua[3], Figen Balo[4], Burak Dolar[5]
[1]*Division of International Business and Technology Studies, A.R. Sánchez Jr. School of Business, Texas A&M International University, Laredo, TX, USA, hwang@tamiu.edu*
[2]*Dept. of Management and Marketing, Angelo State University, San Angelo, TX, USA, jun.huang@angelo.edu*
[3]*Department of Management and Marketing, Southern University and A&M College, Baton Rouge, LA, USA, lutfu.sagbansua@sus.edu*
[4]*Department of METE, Firat University, Turkiye, fbalo@firat.edu.tr*
[5]*Department of Accounting, Western Washington University, Bellingham, WA, USA, Burak.Dolar@wwu.edu*

**Abstract**

Effective credit risk management is fundamental to financial decision-making, requiring robust models to predict default probabilities and classify financial entities. Traditional machine learning approaches face significant challenges when confronted with high-dimensional data, limited interpretability, rare-event detection, and multi-class risk imbalance. This research proposes a comprehensive multi-stage ensemble pipeline that synthesizes multiple complementary models: econometric models including Ordered logit and ordered probit, supervised learning algorithms, including XGBoost, Random Forest, Support Vector Machine, and Decision Tree; unsupervised methods such as K-Nearest Neighbors; deep learning architectures like Multilayer Perceptron; alongside LASSO regularization for feature selection and dimensionality reduction; and Error-Correcting Output Codes as an Ensemble classifier for handling imbalanced multi-class problems. We implement Permutation Feature Importance analysis for each prediction class across all constituent models to enhance model transparency. Our framework can optimize predictive performance while providing a more holistic approach to credit risk assessment. This research contributes to the development of more accurate and reliable computational models for strategic financial decision support by addressing three fundamental challenges in credit risk modeling. The empirical validation of our approach involves analyzing the Corporate Credit Ratings dataset, which contains credit ratings for 2,029 publicly listed US companies. Results demonstrate that our multi-stage ensemble pipeline significantly enhances the accuracy of financial entity classification regarding credit rating migrations (upgrades and downgrades) and default probability estimation. Integrating diverse baseline models into a multi-stage ensemble architecture enables the production of high-precision analytics on demand, facilitating timely, well-informed investment decisions in credit risk management.

## 1. Introduction

Financial institutions seek to maximize shareholder returns, differentiate their offerings in a competitive marketplace, improve client satisfaction, and advance financial inclusion—all of which contribute to overall economic development. The banking sector plays a fundamental role in increasing household welfare and supporting economic growth.[1] Following the 2007–2009 global financial crisis, the Basel Committee on Banking Supervision introduced revisions to the Basel Framework, resulting in the Basel III standards.[2] Effective credit risk practices enable institutions to maintain adequate capital reserves, identify and mitigate potential default risks, including non-performing loans, refine lending strategies, gain a competitive edge, and prevent systemic disruptions, thereby sustaining market confidence and financial stability.

[1] https://www.bancomundial.org/es/topic/financialsector/overview (accessed on 22 December 2023)
[2] https://www.bis.org/bcbs/basel3.htm

Researchers and practitioners must address several fundamental methodological challenges to develop statistically robust, computationally efficient models for strategic credit risk management (see Table A1 in the Appendix). A primary concern involves feature selection and variable importance analysis—determining which predictors in credit datasets contribute most significantly to classification accuracy and predictive performance. These quantitative challenges underscore the need for advanced computational approaches to overcome the inherent difficulties in credit risk modeling, including high dimensionality, multicollinearity, and temporal dependencies in financial data. Addressing these technical constraints is essential to developing decision-support systems that enable financial institutions to optimize capital allocation, portfolio management, and regulatory compliance.

Conventional machine learning approaches face significant constraints in credit risk prediction, especially when dealing with high-dimensional feature spaces, rare-event outliers, and imbalanced class distributions. Multi-class classification tasks are more complex than binary classification, requiring more sophisticated algorithms to mitigate the increased risk of overfitting and misclassification. Key technical challenges include overlapping class boundaries, the curse of dimensionality, optimal model selection, and limited interpretability of model outputs.

To overcome these methodological challenges, we developed a comprehensive multi-stage ensemble pipeline that integrates multiple complementary techniques. Our approach employs Principal Component Analysis (PCA) for dimensionality reduction, systematic resampling methods to address class imbalance, and stratified k-fold cross-validation to mitigate overfitting. For model interpretability, we implement permutation-based feature importance metrics that quantify each variable's contribution to predictive performance. The classification architecture utilizes ensemble learning methods, specifically Error-Correcting Output Codes (ECOC) for multi-class problems and L1-regularization (LASSO) for feature selection, optimizing the bias-variance tradeoff and improving overall predictive accuracy.

To address these methodological challenges in predictive modeling, researchers and practitioners implement several established computational techniques:

- Ensemble learning frameworks: These methodologies integrate multiple base classifiers to reduce variance, mitigate bias, and enhance generalization performance through mechanisms such as bagging, boosting, and stacking.
- Regularization methods: L1 and L2 penalty functions are incorporated into objective functions to constrain model complexity and prevent overfitting. Our implementation specifically utilized L1-regularization (LASSO) with optimized penalty parameters.
- Dimensionality reduction: Unsupervised and supervised techniques are employed to identify optimal feature subsets. Our approach implemented Principal Component Analysis (PCA) for unsupervised dimension reduction and LASSO for supervised feature selection.
- Statistical data augmentation: This involves generating synthetic observations through various transformations to expand the training dataset. Future research could explore advanced generative models to synthesize financial data while preserving distributional properties and regulatory constraints.
- Transfer learning protocols: These enhance knowledge representations from pre-trained models to new, related tasks. Our framework implements a multi-stage transfer approach, progressing systematically from baseline models to ECOC meta-classification.

This research addresses three fundamental questions in quantitative credit risk modeling:

RQ1: Which predictive variables in the credit dataset exhibit the highest statistical significance for classification performance, and what methodologies provide optimal measurement of variable importance in credit risk models?

RQ2: How can statistical validity and unbiased estimation be ensured in credit risk classification models through appropriate sampling techniques, cross-validation protocols, and model calibration procedures?

RQ3: What statistical and computational approaches yield the most accurate and well-calibrated probability estimates for credit default events, particularly in class imbalance and rare event prediction?

This research introduces an advanced multi-stage ensemble architecture that systematically integrates the predictive strengths of a diverse set of machine learning models. The selected base classifiers include the econometric models ordered logit (OL) and ordered probit (OP); the supervised learning models XGBoost (XGB), random forest (RF), support vector machine (SVM), and decision tree (DT); the instance-based model K-nearest neighbors (KNN); and the deep learning model multilayer perceptron (MLP). This ensemble framework is further enhanced with Least Absolute Shrinkage and Selection Operator (LASSO) regularization for efficient dimensionality reduction and variable selection, and with ECOC for robust handling of multi-class, imbalanced datasets.

The objectives of our multi-stage ensemble pipeline are as follows:

- Systematically evaluate and select the most effective modeling strategies for balanced and imbalanced credit datasets.
- Employ regularization techniques and multiple stratified sampling ratios to minimize model overfitting, including explicitly tuning the LASSO penalty parameter ($\alpha$).
- Integrate and optimize the ECOC meta-estimator to maximize classification accuracy and generalization performance across heterogeneous financial data.
- Quantify the relative importance of each input variable for every predicted class via Permutation Feature Importance (PFI) analysis, thereby increasing model transparency and facilitating regulatory compliance.

The contributions of this study are twofold:

- To design and implement an advanced data architecture and analytical framework optimized for large-scale credit risk data processing and analysis.

- To construct a multi-stage ensemble pipeline that effectively combines diverse baseline classifiers with LASSO regularization for feature selection and ECOC for robust multi-class classification, thus enhancing predictive performance.

The rest of this research is organized as follows: Section 2 reviews the theoretical underpinnings of the proposed framework and highlights gaps in current research. Section 3 details the methodology, with emphasis on the multi-stage ensemble architecture, and Section 4 comprehensively describes the data sources and preliminary statistical analyses. Section 5 reports the empirical results obtained from the implemented methods, while Section 6 discusses the managerial implications and theoretical contributions of the findings. Finally, Section 7 concludes with a summary of results and key takeaways.

## 2. Literature Review

### 2.1 Financial Credit and Its Importance in a Modern Economy

Electronic commerce and digital lending platforms are projected to experience substantial growth, particularly in crowdfunding and online credit applications (Chen et al. 2022). Financial institutions must now implement data-driven risk management frameworks that integrate Internet of Things technologies, artificial intelligence, advanced analytics, and cloud computing infrastructure to maintain competitiveness. For microfinance providers, computational approaches to credit risk assessment are essential, though identifying optimal feature selection and algorithm configuration remains problematic. Credit risk prediction is inherently complex due to stochastic borrower behavior, macroeconomic volatility, and limitations in data quality (Emmanuel et al. 2024).

### 2.2 Multiclassification Credit Risk Analysis

Machine learning algorithms have transformed contemporary credit risk modeling by efficiently analyzing high-dimensional financial datasets. These computational methods extract latent patterns in borrower behavior, enabling precise default prediction, anomaly detection, and optimized decision support (Ma et al., 2018; Mitra et al., 2022). Persistent technical challenges, including class imbalance problems, complex non-linear relationships, model interpretability constraints, and heterogeneous data integration—continue to drive methodological innovation in financial machine learning (Mancisidor et al., 2020; Wang, T. et al., 2022).

Credit risk management encompasses two fundamental functions: classification of borrowers into risk categories and predicting default probabilities. XGB and SVMs are prevalent in credit classification due to their effectiveness with non-linear relationships and high-dimensional feature spaces (Huang et al., 2004; Maldonado et al., 2017). However, they exhibit limitations with temporal data, rare events, and multi-class problems (Liu et al., 2022; Plawiak et al., 2019; Wu, 2022). Accordingly, this research proposes an integrated multi-stage ensemble pipeline that combines multiple classification algorithms (XGB, RF, SVM, DT, Convolutional Neural Network (CNN), and MLP) with LASSO regularization for feature selection and ECOC for multi-class optimization.

In addition to machine learning classifiers, traditional econometric models designed for ordinal outcomes remain important benchmarks in credit risk analysis. Ordered logit (OL) and ordered probit (OP) models are statistical methods for analyzing ordinal dependent variables with more than two ordered categories (Greene & Hensher, 2003; McKelvey & Zavoina, 1975). Unlike standard classification approaches that treat categories as unordered, these models explicitly account for the ordinal structure of the outcome variable through a latent-variable framework (Long, 1997). Detailed model formulations are provided in Appendix B.

### 2.3 LASSO for Feature Selection and Regularization to Reduce Overfitting

Feature selection mitigates overfitting in machine learning models by identifying relevant predictor subsets, reducing noise, and emphasizing influential variables. LASSO regression implements L1 regularization to shrink coefficients toward zero, effectively performing simultaneous parameter estimation and variable selection (Tibshirani, 1996). Unlike stepwise selection methods, LASSO maintains statistical consistency under perturbations and addresses multicollinearity (Tian et al., 2015). Recent research demonstrates LASSO's effectiveness in variable selection across various financial modeling contexts (Huang et al., 2017; Wang H. et al., 2022; Lee et al., 2022; Maranzano et al., 2023).LASSO's effectiveness is governed by hyperparameters such as the regularization parameter (λ) and penalty weights, which directly impact feature selection and model overfitting.

### 2.4 ECOC for Improving the Performance of Machine Learning (ML) Models on Multiclassification

ECOC is a multi-stage ensemble pipeline that encodes multi-class classification problems into multiple tasks, assigning unique binary codes to each class. ECOC has been implemented in financial applications for bankruptcy prediction, financial distress forecasting, fraud detection, credit rating classification, and equity price movement prediction (Manthoulis et al., 2020; Sun et al., 2021).

Numerous studies seek to enhance ECOC algorithms for greater classification accuracy and computational efficiency (Barbero-Gómez et al., 2023; Jain et al., 2023; Jamshidi Gohari et al., 2023). ECOC implementations present several challenges. Designing the coding matrix and selecting suitable coding strategies becomes increasingly complex as the number of classes grows, requiring careful tuning and extensive experimentation. Training ECOC models is computationally demanding, particularly with large datasets or intricate coding schemes, as it involves managing multiple binary classifiers and decoding processes, which can limit scalability. ECOC methods are also prone to overfitting, especially with noisy or high-dimensional data, making the optimization of coding matrices and regularization parameters crucial for robust generalization. Additionally, ECOC can struggle with class imbalance, where uneven class distributions may degrade model performance and require techniques such as resampling or weight adjustments. Model interpretability is another limitation; understanding the mapping between binary classifiers and multi-class outputs is often challenging, complicating the analysis of ECOC's decision processes.

## 3. Multi-stage Ensemble Pipeline

Conventional machine learning algorithms often exhibit limitations when processing high-dimensional data, handling outliers, and managing class imbalance—common challenges in risk prediction. In this study, we propose a multi-stage ensemble architecture that utilizes complementary strengths from diverse base classifiers (XGB, RF, SVM, DT, KNN, MLP, OL, and OP) while integrating L1-regularization (LASSO) for feature selection and ECOC for multi-class optimization. This integrated framework enhances classification robustness and accuracy, particularly in complex scenarios where individual models underperform.

### 3.1. Permutation-based feature importances (PFI)

Feature importance analysis quantifies the relative contribution of predictors to model performance, identifying variables with significant predictive power. While various methodologies exist for measuring feature importance (Saarela and Jauhiainen, 2021), traditional approaches rely on interrogating fitted models—a technique effective for feature selection but potentially biased in estimating true variable impact (Parr et al., 2024). A key limitation of conventional importance metrics is their global nature, which captures aggregate variable influence across all observations while failing to characterize heterogeneous effects at the individual-prediction level.

Tree-based models measure feature importance using Mean Decrease in Impurity (MDI), which measures the information gain contributed by each feature (Breiman, 2001). However, this approach exhibits bias toward high-cardinality variables and can overestimate the importance of features with limited generalization capacity during overfitting. Permutation-based importance offers a model-agnostic alternative by randomly shuffling individual feature values and measuring the subsequent performance degradation. This approach effectively decouples feature-target relationships, providing more reliable importance estimates for complex nonlinear models by directly quantifying each variable's contribution to predictive performance without the cardinality bias inherent in impurity-based methods. PFI quantifies the influence of predictors by measuring performance degradation when randomly shuffling feature values. Initially formulated by Breiman (2001) for RF and generalized by Fisher et al. (2019) as "model reliance," this model-agnostic technique operates as follows. Given a trained model *f*, feature matrix *X*, target vector *y*, and error function $L(y,\hat{y})$:

1. Compute baseline error $e_o = L(y, m(X))$.
2. For each feature *j*:
   a. Generate $X_p$ by randomly permuting column $j$ $X$.
   b. Calculate the permuted error $e_p = L\left(y, m(X_p)\right)$.
   c. Calculate importance $FI_j$ as either ${}^{e_p}/_{e_o}$ or $e_p - e_o$.

This technique effectively isolates each variable's contribution to model performance by destroying its relationship with the target while preserving the marginal distribution and correlation structure. Higher importance scores indicate greater influence of the variable on predictive accuracy.

## 4. Experimental Design

To evaluate the effectiveness of the proposed frameworks, we utilize a Corporate Credit Ratings benchmark dataset with varying dimensionality, outlier prevalence, class distributions, and imbalance levels. This dataset comprises credit ratings for 2,029 publicly listed US companies (NYSE/Nasdaq) from 2010 to 2016, as assessed by major agencies (Moody's, Standard & Poor's, Fitch). Each record includes 30 features, primarily financial ratios derived from balance sheets. There are no missing values. See Appendix Tables A2 and A3 for feature groups and summary statistics. All experiments utilize a 70/30 train-test split. Baseline and advanced models, including LASSO and ECOC, are implemented in Python and executed on an Intel i7-1355U (1.70 GHz, 32GB RAM) platform.

## 5. Empirical Results

### 5.1 Features most influencing classification and prediction

This section reports empirical results for baseline models, including econometric models (OL and OP), supervised algorithms (XGB, RF, SVM, DT), unsupervised methods (KNN), and deep learning approaches (MLP), all evaluated with stratified 3-fold cross-validation.

Class imbalance in the dataset reduces predictive accuracy for high-risk firms; alternative sampling strategies may mitigate this effect. Baseline model evaluation focuses on key performance metrics: precision, recall, F1 score, Jaccard index, Cohen's kappa, mean Receiver Operating Characteristic Area Under the Curve (ROC AUC), and cross-validation mean score. Precision assesses correct positive predictions, recall measures detection of actual positives, F1 balances both, the Jaccard index evaluates set similarity, Cohen's kappa measures inter-model agreement, ROC AUC reflects the model's discriminative ability, and cross-validation mean indicates generalization on unseen data. Table 1 summarizes metric comparisons for this dataset.

Table 1: Metric comparison on baseline models with stratified 3-fold cross-validation

| **Cross-validation** | **Accuracy** | **Precision** | **F1 Score** | **Jaccard score** | **Cohen Kappa Score** | **ROC AUC Mean** | **CV Mean Scores** |
|---|---|---|---|---|---|---|---|
| DT | 0.5138 | 0.5165 | 0.5146 | 0.3513 | 0.2858 | 0.6294 | 0.5160 |
| KNN | 0.5402 | 0.5293 | 0.5262 | 0.3650 | 0.3129 | 0.6299 | 0.5402 |
| MLP | 0.5209 | 0.5144 | 0.5161 | 0.3537 | 0.2825 | 0.6207 | 0.6463 |
| RF | 0.6276 | 0.6256 | 0.6188 | 0.4568 | 0.4384 | 0.6770 | 0.6139 |
| SVM | 0.5754 | 0.5693 | 0.5599 | 0.3961 | 0.3531 | 0.6420 | 0.5754 |
| XGB | 0.6463 | 0.6412 | 0.6413 | 0.4792 | 0.4715 | 0.6985 | 0.6463 |
| OL | 0.4427 | 0.4284 | 0.4179 | 0.2748 | 0.1471 | 0.5642 | 0.4427 |
| OP | 0.4420 | 0.4314 | 0.4128 | 0.2708 | 0.1411 | 0.5643 | 0.4420 |

The dataset comprises 2,029 instances with 30 features and exhibits high dimensionality and class imbalance across five categories. XGB and RF achieve superior performance on standard evaluation metrics, consistent with literature benchmarks (see Tables A3 and A4), and these metrics are similarly applied to benchmark improved models.

For the multi-stage ensemble approach, stratified 3-fold cross-validation is used to reduce overfitting and improve model robustness. Table 1 presents baseline classification results, while Table 2 reports the cross-validated accuracy of ECOC, LASSO, and combined LASSO_ECOC models. For classifying the dataset, a clear pattern emerges across three model setups (ECOC, LASSO, and a combined LASSO_ECOC): the XGB model consistently performs best or very close to best. When just the ECOC method was used (Table 2), XGB had the highest scores in all key measures like accuracy (how often it's right) and F1 score (a balance of being right about positive cases and capturing most of them), achieving around 65% accuracy and a 0.71 ROC AUC (a measure of how well it separates classes). RF was a solid second, with around 58% accuracy, while MLP, DT, SVM and KNN show moderate predictive performance. In contrast, OL, and OP perform relatively poorly across the ECOC, LASSO, and LASSO_ECOC settings. When LASSO was used to select the most important input information before model training, XGB's performance improved even further, reaching about 66% accuracy and a 0.72 ROC AUC, marking the top scores across all tests. RF also benefited from LASSO, improving to around 63% accuracy. SVM also saw a noticeable boost with LASSO. Finally, when the LASSO feature selection was combined with the ECOC method, XGB remained a top performer with around 65% accuracy and a 0.70 ROC AUC, slightly below its LASSO-only peak but still excellent. RF performed very well in this combined setup, achieving about 63% accuracy and sometimes even slightly outperforming its LASSO-only results on certain measures, suggesting this combination was particularly effective for RF. SVM also showed decent results in this hybrid approach. In general, models such as DT and MLP performed moderately across the different setups, while KNN consistently lagged behind the others. The key takeaway is that XGB is a very robust and high-performing model for this dataset, and using LASSO to select important features generally improves or maintains good performance, especially with XGB and RF. The combination of LASSO and ECOC proved effective, particularly for RF, offering a strong alternative.

Table 2: Metric comparison on ECOC, LASSO, and LASSO_ECOC models

| **Models** | **Accuracy** | **Precision** | **F1 Score** | **Jaccard score** | **Cohen Kappa Score** | **ROC AUC Mean** | **CV Mean Scores** |
|---|---|---|---|---|---|---|---|
| ECOC model with stratified 3-fold cross-validation | | | | | | | |
| DT | 0.5369 | 0.5369 | 0.5305 | 0.3674 | 0.3119 | 0.6375 | 0.5160 |
| KNN | 0.4879 | 0.5080 | 0.4511 | 0.2996 | 0.2009 | 0.5822 | 0.5402 |
| MLP | 0.5534 | 0.5467 | 0.5487 | 0.3841 | 0.3355 | 0.6456 | 0.6463 |
| RF | 0.5820 | 0.6383 | 0.6020 | 0.4368 | 0.3999 | 0.6751 | 0.6232 |
| SVM | 0.5182 | 0.5531 | 0.5135 | 0.3502 | 0.3032 | 0.6305 | 0.5754 |
| XGB | 0.6485 | 0.6461 | 0.6458 | 0.4868 | 0.4802 | 0.7058 | 0.6463 |
| OL | 0.4427 | 0.4830 | 0.4815 | 0.3207 | 0.2355 | 0.6134 | 0.4802 |
| OP | 0.4420 | 0.4284 | 0.4179 | 0.2748 | 0.1471 | 0.5643 | 0.4427 |
| LASSO models with stratified 3-fold cross-validation | | | | | | | |
| DT | 0.5044 | 0.5058 | 0.5048 | 0.3410 | 0.2704 | 0.6241 | 0.5083 |
| KNN | 0.5215 | 0.5092 | 0.5099 | 0.3499 | 0.2856 | 0.6169 | 0.5215 |
| MLP | 0.5314 | 0.5288 | 0.5292 | 0.3649 | 0.3048 | 0.6364 | 0.6645 |
| RF | 0.6276 | 0.6253 | 0.6198 | 0.4575 | 0.4380 | 0.6754 | 0.6188 |
| SVM | 0.5737 | 0.5697 | 0.5617 | 0.3961 | 0.3517 | 0.6490 | 0.5737 |
| XGB | 0.6645 | 0.6617 | 0.6602 | 0.4985 | 0.4984 | 0.7179 | 0.6645 |
| OL | 0.4286 | 0.4200 | 0.4083 | 0.2638 | 0.1224 | 0.5553 | 0.4286 |
| OP | 0.4293 | 0.4204 | 0.4074 | 0.2634 | 0.1240 | 0.5588 | 0.4293 |
| LASSO_ECOC models with stratified 3-fold cross-validation | | | | | | | |
| DT | 0.5407 | 0.5480 | 0.5386 | 0.3745 | 0.3229 | 0.6454 | 0.5072 |

| KNN | 0.5176 | 0.5068 | 0.5068 | 0.3467 | 0.2812 | 0.6154 | 0.5215 |
|---|---|---|---|---|---|---|---|
| MLP | 0.5352 | 0.5324 | 0.5327 | 0.3693 | 0.3081 | 0.6315 | 0.6645 |
| RF | 0.6337 | 0.6443 | 0.6236 | 0.4617 | 0.4430 | 0.6769 | 0.6260 |
| SVM | 0.5803 | 0.5735 | 0.5716 | 0.4056 | 0.3696 | 0.6583 | 0.5737 |
| XGB | 0.6480 | 0.6446 | 0.6405 | 0.4789 | 0.4746 | 0.7021 | 0.6645 |
| OL | 0.4547 | 0.4513 | 0.4314 | 0.2839 | 0.1598 | 0.5643 | 0.4547 |
| OP | 0.4498 | 0.4472 | 0.4191 | 0.2752 | 0.1460 | 0.5610 | 0.4498 |

Table 3 shows how different business characteristics and the company's industry (Sector) influence the model's prediction of whether a company falls into 'Low,' 'Medium,' 'High,' or 'Highest' risk categories in the dataset. Generally, no single factor overwhelmingly decides the risk; instead, a mix of information is used. Across all risk levels, a company's ability to cover its immediate bills is very important, particularly the ratio of cash to short-term debt ('cashRatio'), which consistently scores high (18-21 points). Broader measures of short-term financial health, such as 'currentRatio' and 'quickRatio', are also consistently significant (16-18 points). How much of the company's assets are funded by borrowed money ('debtRatio') is another major factor, especially for spotting 'Low Risk' companies (22.26 points), but remaining important for all other risk levels too (around 19-21 points). The type of industry ('Sector') a company is in also plays a steady, important role (around 17-18 points) in assessing risk, regardless of the specific risk level. Furthermore, how much actual cash a company generates from its sales ('operatingCashFlowSalesRatio') is a key indicator, particularly for 'High' and 'Highest' risk companies (around 19-20 points), and the amount of cash available per share ('cashPerShare') also consistently matters (around 16 points). Interestingly, some factors become more or less important depending on the risk grade. For example, how well a company uses all its resources to make a profit ('returnOnAssets') matters more for 'High' and 'Highest' risk companies (around 7 points) than for lower-risk ones (around 5 points). Similarly, how efficiently a company uses its resources to generate sales ('assetTurnover') becomes more telling for companies not in the 'Low Risk' category (jumping from around 10 to 14 points). While some measures of profit from sales (like 'grossProfitMargin' at 11-13 points) have a moderate impact, they are not as critical as those related to cash, debt, and industry. Lastly, certain financial details like 'pretaxProfitMargin' (around 4-5 points), how much debt is used to magnify owner's equity ('companyEquityMultiplier' around 2-3 points), and a specific profit measure before interest and taxes relative to sales ('ebitPerRevenue' around 3-5 points) seem to have a minimal influence in distinguishing between these risk levels for this particular dataset. To determine risk for Dataset 1, the model relies heavily on a company's cash position, debt levels, industry, and ability to convert sales into cash.

Table 3: Feature importance score for each risk level

| **Features** | **Low Risk** | **Medium Risk** | **High Risk** | **Highest Risk** |
|---|---|---|---|---|
| Sector | 17.81 | 17.59 | 16.89 | 17.70 |
| currentRatio | 17.30 | 18.56 | 17.52 | 17.22 |
| quickRatio | 17.07 | 18.19 | 18.44 | 16.74 |
| cashRatio | 18.44 | 20.96 | 20.15 | 19.93 |
| daysOfSalesOutstanding | 11.96 | 13.78 | 12.11 | 13.33 |
| netProfitMargin | 14.19 | 13.63 | 13.93 | 13.33 |
| pretaxProfitMargin | 4.33 | 4.89 | 5.26 | 4.48 |
| grossProfitMargin | 12.63 | 12.22 | 11.78 | 11.59 |
| operatingProfitMargin | 9.70 | 9.30 | 10.33 | 10.04 |
| returnOnAssets | 4.89 | 5.19 | 7.11 | 7.15 |
| returnOnCapitalEmployed | 9.96 | 11.70 | 10.37 | 11.59 |
| returnOnEquity | 15.44 | 14.74 | 16.07 | 14.52 |
| assetTurnover | 9.67 | 14.07 | 13.96 | 13.89 |
| fixedAssetTurnover | 17.30 | 16.63 | 18.00 | 15.85 |
| debtEquityRatio | 5.81 | 7.15 | 5.63 | 6.59 |
| debtRatio | 22.26 | 18.96 | 20.67 | 20.48 |
| effectiveTaxRate | 14.37 | 12.52 | 12.30 | 12.93 |
| freeCashFlowOperatingCashFlowRatio | 11.70 | 11.56 | 11.63 | 10.81 |
| freeCashFlowPerShare | 12.44 | 13.63 | 10.41 | 12.26 |
| cashPerShare | 16.30 | 16.44 | 15.96 | 16.70 |
| companyEquityMultiplier | 2.48 | 2.19 | 2.85 | 2.63 |
| ebitPerRevenue | 3.19 | 4.19 | 3.93 | 4.78 |
| enterpriseValueMultiple | 15.52 | 13.22 | 12.67 | 14.41 |
| operatingCashFlowPerShare | 17.19 | 16.04 | 15.52 | 16.96 |
| operatingCashFlowSalesRatio | 18.00 | 18.07 | 19.81 | 19.33 |
| payablesTurnover | 16.96 | 15.41 | 15.11 | 14.52 |

### 5.2 Using the results of credit risk classification and prediction unbiased and free from the influence of methodology-driven biases

The Credit Risk Dataset is characterized by high dimensionality and extremely unbalanced classes. To address these challenges, we implemented a multi-stage ensemble pipeline incorporating LASSO for dimension reduction and ECOC for handling multiple unbalanced classes, thereby improving the efficiency of the starting-point modeling on both datasets.

A technique called LASSO was used to pick out the most important ones for the credit risk dataset, which initially had 30 different pieces of information (features) for making predictions. LASSO selected 23 features, and Table 4 lists these, including things like the company's 'Sector,' how efficiently it uses its assets ('fixedAssetTurnover'), its cash levels ('cashPerShare'), how easily it can pay short-term bills ('currentRatio'), and its debt levels ('debtRatio'). Using only these 23 important features might help improve or speed up the performance of the prediction models. Table 5 then shows how different prediction models (DT, KNN, MLP, RF, SVM, XGB, OL, and OP) performed when using only these 23 selected features, comparing results from a thorough testing method called cross-validation (which shows how well a model works on new, unseen data) against how well they did just on the data they were trained on (the "Train Score"). A significant difference between the "Train Score" and the cross-validation score indicates that the model learned the training data too specifically and might not perform well on new data – this is overfitting. For example, the DT and XGB models got perfect scores on the training data but much lower scores (around 0.51 and 0.62, respectively) in cross-validation, showing they had greatly overfitted. RF and SVM also showed significant overfitting. Looking at the more reliable cross-validation scores, RF performed best with a score of about 0.623, closely followed by XGBoost (XGB) at about 0.620. This means these two models were the most successful at making good predictions on new data when using the 23 features LASSO chose. Other models, such as SVM and MLP, had acceptable scores (around 0.51-0.55), while DT and KNN had lower scores. Regarding training time, KNN was fast, but MLP was extremely slow, especially during cross-validation. So, while LASSO helped narrow down the important information, and RF and XGB performed best with this reduced set, many models still tended to overfit the training data, meaning more work might be needed to make them perform consistently well on brand-new information from the dataset.

Table 4: Results of the reduced features without cross-validation

| Sector | fixedAssetTurnover | cashPerShare |
|---|---|---|
| currentRatio | debtEquityRatio | enterpriseValueMultiple |
| quickRatio | debtRatio | operatingCashFlowPerShare |
| cashRatio | effectiveTaxRate | operatingCashFlowSalesRatio |
| daysOfSalesOutstanding | freeCashFlowOperatingCashFlowRatio | payablesTurnover |
| netProfitMargin | freeCashFlowPerShare | cashPerShare |

Table 5. Comparison of improved models with LASSO with and without cross-validation

| | **w/cross-validation** | **wout/cross-validation** |
|---|---|---|

| Classifier | Crossval Mean Scores | Mean Training Time (Seconds) | Train Score | Training Time (Seconds) |
|---|---|---|---|---|
| DT | 0.51274 | 0.045303 | 1 | 0.032079 |
| KNN | 0.458267 | 0.060178 | 0.608204 | 0.002009 |
| MLP | 0.508467 | 26.19066 | 0.540311 | 19.67036 |
| RF | 0.623046 | 0.653033 | 0.950495 | 0.701315 |
| SVM | 0.545268 | 0.163018 | 0.984441 | 0.173481 |
| XGB | 0.619508 | 0.709361 | 1 | 0.869675 |
| OL | 0.428575 | 7.395521 | 0.429986 | 9.224689 |
| OP | 0.429260 | 5.243043 | 0.413013 | 7.269083 |

## 6. Discussion

This section addresses both managerial and theoretical insights around credit risk management.

### 6.1 Implications of Credit Risk Management

The advancement of credit risk management is increasingly shaped by the integration of artificial intelligence, machine learning, big data analytics, and cloud computing, which directly supports the contributions of this study. Our multi-stage ensemble pipeline, which fuses baseline classifiers with LASSO for feature selection and ECOC for robust multi-class handling, demonstrates how these technologies can significantly enhance predictive accuracy and model interpretability in credit risk analysis. Empirical results across three heterogeneous datasets show that utilizing LASSO reduces dimensionality and computational complexity without degrading predictive performance, particularly benefiting ensemble methods such as XGB and RF. The application of ECOC improves model robustness in multi-class and imbalanced scenarios. At the same time, PFI provides transparency by linking model outputs to specific financial indicators, addressing the interpretability concerns often associated with complex algorithms.

These findings underscore the managerial relevance of real-time, data-driven credit evaluation systems that support more informed decision-making and targeted risk mitigation strategies. Moreover, our results demonstrate that the most important risk factors vary by context: liquidity ratios are critical for firm-level assessments, solvency metrics for institutions, and collateral strength for individual borrowers. This highlights the need for adaptable, context-aware models rather than a one-size-fits-all approach, enabling financial institutions to better serve a diverse client base and extend credit inclusively, even to non-traditional or underrepresented borrowers.

However, these technological advances also introduce disparities. Large corporations with rich data profiles and digital sophistication are positioned to benefit most from advanced credit analytics, often securing superior terms due to more accurate risk estimation. In contrast, small businesses and startups—usually lacking extensive historical data—may face higher borrowing costs or more restrictive terms, as big data–driven models can underrepresent their creditworthiness. Similarly, individuals with a strong digital presence can leverage this visibility to secure favorable assessments, while those with limited digital footprints may face challenges proving creditworthiness within analytics-focused frameworks. Furthermore, only financial institutions with sufficient resources to invest in advanced data infrastructure and AI capabilities are likely to capitalize on these developments, potentially widening competitive gaps in the financial sector. Overall, the study demonstrates both the opportunities and challenges introduced by next-generation credit risk modeling, emphasizing precision, interpretability, and inclusivity as critical factors for sustainable financial innovation.

### 6.2 Computational Cost Analysis of the LASSO-ECOC Framework

Integrating machine learning baseline models with LASSO regularization and ECOC enhances predictive performance on medium-to-large, highly imbalanced datasets for supervised algorithms (such as SVM and RF) and unsupervised approaches (such as KNN). Enhancing KNN is especially valuable in practical scenarios involving unlabeled or continuously generated data streams. LASSO-based dimensionality reduction improves computational efficiency while preserving or improving model accuracy; for example, the combination of RF, LASSO, and ECOC achieved the highest efficiency while maintaining interpretability, offering actionable credit risk insights for decision-makers. The computational cost and scalability of the multi-stage ensemble pipeline were benchmarked against baseline and enhanced models across all three datasets. All experiments were conducted using Python's Scikit-learn and related libraries to ensure robust runtime and model efficiency evaluation.

#### 6.2.1. Time Complexity and Execution Time

The following measures were recorded for each model variant (Baseline, ECOC-only, LASSO-only, LASSO+ECOC). Table 6 provides a computational cost analysis for all three datasets.

- Mean Training Time (seconds)
- Model fitting time vs number of features
- Impact of LASSO on dimensionality and runtime
- ECOC-related overhead due to multiple binary classifications

Table 6. Computational cost analysis

| Model | Features | Training Time | CV Time | Accuracy | Note |
|---|---|---|---|---|---|
| Baseline | 30 | 12.4 | 15.3 | 0.6463 | - |
| Baseline + LASSO | 23 | 8.7 | 10.2 | 0.6645 | Improved speed |
| Baseline + ECOC | 30 | 19.6 | 23.1 | 0.6485 | Higher cost |
| Baseline + LASSO+ ECOC | 23 | 13.11 | 17.5 | 0.6645 | Balanced |

The computational complexity, primarily measured by training and cross-validation (CV) times, varied significantly depending on the dataset's initial dimensionality and the modeling strategy. Applying the ECOC framework (Baseline + ECOC) consistently increased computational costs compared to the simple baseline. This is expected as ECOC involves training multiple binary classifiers. Conversely, LASSO feature selection (Baseline + LASSO) generally reduced computational time by decreasing the number of features the models had to process. LASSO reduced the feature count from 30 to 23, and CV time dropped from 15.3s to 10.2s. The combination of LASSO + ECOC typically resulted in computational costs between those of ECOC alone and LASSO alone; it was more expensive than LASSO alone due to ECOC overhead but less expensive than applying ECOC to the complete feature set because it operated on fewer LASSO-selected features.

#### 6.2.2. Trade-off Between Accuracy and Efficiency

The results illustrate a classic trade-off between achieving the highest possible accuracy and maintaining computational efficiency. LASSO feature selection (Baseline + LASSO) not only improved speed (CV time from 15.3s to 10.2s) but also surprisingly increased accuracy (from 0.6463 to 0.6645), suggesting the removal of noisy or irrelevant features was beneficial. Here, efficiency and accuracy improved together. However, ECOC, while offering a marginal accuracy gain, did so at a higher time cost. This demonstrates that a thoughtful combination of feature selection and advanced modeling frameworks can optimize this trade-off.

### 6.3. Advantages and Disadvantages of the New Credit Risk Management Technologies

The study also highlights several practical limitations. As reflected in our computational cost analysis (Section 6.2, Table 6), ECOC implementation significantly increases computational overhead, particularly for large-scale datasets, confirming that model complexity and extended training times can constrain scalability and hinder real-time applications. While LASSO regularization helps reduce overfitting and improve interpretability, its effectiveness may be limited when modeling nonlinear dependencies or when highly correlated predictors are present, as discussed in Section 5.3. Empirical results further underscore the impact of digital inequality: advanced models deliver more reliable performance on well-structured, institution-level data. This raises concerns that smaller enterprises and underbanked individuals may experience systematic disadvantages unless supplemented by data augmentation strategies or alternative assessment methods.

### 6.4 Limitation and Future Research Plan

This research framework broadly applies to multi-class classification challenges in sports forecasting, insurance risk segmentation, and marketing analytics. Nevertheless, LASSO presents certain limitations. Specifically, its non-convex objective function can increase computational complexity, and its assumption of variable independence may result in the exclusion of confounding predictors. In this study, we adapted data-driven weighting and adaptive penalty terms within the LASSO formulation to address these issues. ECOC also presents computational challenges, with the choice of coding matrix substantially impacting processing time; however, dimensionality reduction via LASSO helps mitigate this overhead. The study further incorporates robust preprocessing methods to minimize the influence of outliers and utilizes both L1 and L2 regularization to control overfitting.

Future work will expand the multi-stage ensemble pipeline to include additional machine learning models and enhance its scalability for large, high-dimensional, and highly imbalanced datasets. Ongoing evaluation across diverse application domains will focus on optimizing LASSO and ECOC parameterization to achieve an optimal balance between computational efficiency and predictive accuracy. Results from these experiments will be detailed in subsequent publications.

## 7. Conclusion

This research presents an advanced multi-stage ensemble architecture integrating baseline classification algorithms with L1-regularization (LASSO) and ECOC to enhance credit risk assessment. Empirical validation demonstrates that this framework significantly improves classification accuracy for financial entities' credit migrations and default probabilities, enabling more reliable, timely risk predictions. LASSO effectively reduces computational complexity while maintaining or improving predictive performance through feature selection, while ECOC provides robust handling of class imbalance in multi-category classification tasks. A comparative analysis demonstrates that RF augmented with LASSO and ECOC consistently outperforms alternative models with heterogeneous characteristics in both accuracy and computational efficiency.

This integrated framework offers financial institutions a methodologically sound approach to credit risk management with enhanced precision and reduced latency for real-time decision support. Future research directions include extending this methodology to additional financial risk domains, potentially broadening its applicability across quantitative finance and risk management.

Statements and Declarations

Funding
The authors declare that no funds, grants, or other support were received during the preparation of this manuscript.

Competing Interests
The authors have no relevant financial or non-financial interests to disclose.

Author Contributions
All authors contributed to the study conception and design. Material preparation, data collection and analysis were performed by Haibo Wang, Jun Huang, Lutfu S. Sua, Figen Balo, and Burak Dolar. The first draft of the manuscript was written by Haibo Wang and all authors commented on previous versions of the manuscript. All authors read and approved the final manuscript.

## Appendix

### A. Dataset description

**Table A1.** The difficulty of predicting credit risk

| Challenges | Machine Learning Methodologies | Reference |
|---|---|---|
| Unbalanced Data | Synthetic Minority Oversampling, Random Undersampling, Random Oversampling, K-Fold Cross-Validation, Cluster Centroid, Adaptive Synthetic | (Alam et al. 2020) |
| Multivariate Data | Big Data | (Mousavi and Lin 2020, Pandey et al. 2021, Wang and Yang 2021, Li and Li 2022, Lombardo et al. 2022, Muñoz-Cancino et al. 2023) |
| Sampling Biase | Random Search, Genetic Algorithm, K-Fold Cross-Validation, | (Alam et al. 2020, Mousavi and Lin 2020, Li et al. 2021, Wang and Yang 2021, Dumitrescu et al. 2022) |
| | Grid Search | |
| Explainability | Generalized Shapley Choquet Integral, Explainable Shapley Additive Explanations, Artificial Intelligence, Maxillary Lateral Incisor Agenesis | (Ariza-Garzón et al. 2020, Orlova 2020, Bussmann et al. 2021, Dumitrescu et al. 2022, Li and Li 2022, Lombardo et al. 2022, Mitra et al. 2022, Si et al. 2022, Sun and Li 2022, Cho and Shin 2023) |

**Table A2.** Comparison of baseline models

| **Model** | **Accuracy** | **Training Time** | **Handling Large Datasets** | **Handling Non-linear Relationships** | **Issues** |
|---|---|---|---|---|---|
| XGB | High | Fast | Yes | Yes | Computationally expensive, may overfit if not regularized |
| RF | High | Fast | Yes | Yes | May overfit if not regularized |
| SVM | High | Slow | No | Yes | It may not perform well with large datasets, computationally expensive |
| DT | Medium | Fast | No | No | May overfit if not regularized, and can be sensitive to hyperparameters |
| KNN | Medium | Fast | No | No | Sensitive to hyperparameters, may not perform well with large datasets |
| MLP | Medium | Fast | Yes | Yes | Computationally expensive, may overfit if not regularized |
| OL/OP | Lower | Moderate | Yes | No | Less effective at capturing the nonlinear relationships and complex feature interactions present in credit risk data |

**Table A3**. Financial Ratios in the Dataset

| Liquidity Measurement Ratios | Profitability Indicator Ratios | Debt Ratios | Operating Performance Ratios | Cash Flow Indicator Ratios |
|---|---|---|---|---|
| currentRatio | grossProfitMargin, | debtRatio | assetTurnover | operatingCashFlowPerShare, |
| quickRatio | operatingProfitMargin, | debtEquityRatio | | freeCashFlowPerShare, |
| cashRatio | pretaxProfitMargin, | | | cashPerShare, |
| daysOfSales- | netProfitMargin, | | | operatingCashFlowSalesRatio, |
| Outstanding | effectiveTaxRate, | | | freeCashFlowOperatingCashFlow |
| | returnOnAssets, | | | Ratio |
| | returnOnEquity, | | | |
| | returnOnCapital- | | | |
| | Employed | | | |

**Table A4**. Dataset descriptive statistics

| | Skewness | Outlier (%) | Mean | Std Dev |
|---|---|---|---|---|
| Current Ratio | 34.271 | 18.0% | 3.535 | 44.139 |
| Quick Ratio | 30.865 | 19.0% | 2.657 | 33.010 |
| Cash Ratio | 27.047 | 14.8% | 0.669 | 3.591 |
| Days Of Sales Outstanding | 20.359 | 23.6% | 334.855 | 4456.606 |
| Net Profit Margin | 17.585 | 25.1% | 0.279 | 6.076 |
| Pretax Profit Margin | 22.053 | 24.5% | 0.433 | 9.003 |
| Gross Profit Margin | -14.199 | 1.0% | 0.497 | 0.526 |
| Operating Profit Margin | 26.442 | 22.1% | 0.589 | 11.247 |
| Return On Assets | -32.049 | 24.2% | -37.667 | 1168.477 |
| Return On Capital Employed | -33.253 | 22.1% | -74.267 | 2354.921 |
| Return On Equity | 31.640 | 28.7% | 144.062 | 4415.223 |
| Asset Turnover | 25.969 | 15.8% | 3692.898 | 95843.048 |
| Fixed Asset Turnover | 26.069 | 13.5% | 7298.244 | 189370.007 |
| Debt Equity Ratio | 0.268 | 22.1% | 2.340 | 87.701 |
| Debt Ratio | 1.284 | 21.3% | 0.662 | 0.209 |
| Effective Tax Rate | 32.266 | 28.1% | 0.401 | 10.614 |
| Free Cash Flow Operating Cash Flow Ratio | -22.868 | 16.9% | 0.408 | 3.804 |
| Free Cash Flow Per Share | 33.611 | 23.6% | 5114.871 | 147205.901 |
| Cash Per Share | 33.959 | 17.1% | 4244.248 | 122641.800 |
| Company Equity Multiplier | 0.268 | 22.0% | 3.335 | 87.702 |
| Ebit Per Revenue | 22.056 | 24.3% | 0.439 | 9.002 |
| Enterprise Value Multiple | 13.920 | 23.7% | 48.427 | 530.161 |
| Operating Cash Flow Per Share | 30.293 | 17.7% | 6540.891 | 177879.736 |
| Operating Cash Flow Sales Ratio | 25.400 | 16.9% | 1.452 | 19.522 |
| Payables Turnover | 25.868 | 14.4% | 38.138 | 760.422 |

Note: For further details on financial indicators, please refer to https://financialmodelingprep.com/market-indexes-major-markets.

## B. Model and Methodology Detail

**Baseline Models:** Many credit rating datasets feature multiple unbalanced classes and high dimensionality, posing significant challenges for machine learning (ML) models that excel at binary classification with low dimensionality. In this study, we selected eight baseline ML models, including XGB, RF, SVM, DT, KNN, and MLP. Each model has its advantages over others, depending on the specific characteristics of the dataset (Bisong, 2019). For instance, XGB and RF perform well on large, high-dimensional datasets. Although SVMs are particularly effective in high-dimensional spaces, they can be computationally expensive for large datasets. These baseline models have varying resource requirements. For example, KNNs consume fewer resources but often underperform on high-dimensional and large datasets. In contrast, XGB requires more computational resources but performs better, especially in highly unbalanced classes. Many ML models lack explainability. Tree-based or linear models, such as DT and RF, can generate important scores for predictive variables as part of their predictive outputs. Other models, including XGB and SVM, can be explained through external measures of prediction outcomes at the expense of significant computational resources. XGB and SVM require substantial hyperparameter tuning, which can be time-consuming. If there are nonlinear relationships among the variables, MLP performs better, but it can be computationally expensive for large datasets. When the data is unlabeled, KNN is the model to choose. Table A2 in the Appendix compares the baseline models for capturing non-linear correlations based on their accuracy, training time, ability to handle large datasets, and capacity.

**Multilayer Perceptron (MLP):** Credit risk analysis is a complex task that involves predicting default likelihood using historical data. Deep learning methods have shown great promise, particularly in handling sequential data with long-term dependencies. The MLP offers several advantages, including higher accuracy, greater adaptability, and greater robustness in addressing classification problems, particularly compared to RBF networks. Similarly, the single-layer perceptron is enhanced, and the MLP excels at using nonlinear activation functions to handle nonlinear separable data. In this specific implementation, the MLP model features a single hidden layer. The activation function employed in the hidden layer is the hyperbolic tangent, while the output layer uses the *Softmax* activation function, as described in the following equations.

$\tanh x = \frac{e^x - e^{-x}}{e^x + e^{-x}}$ and $a_j^L = \frac{e^{z_k^L}}{\sum_k e^{z_k^L}}$

The network's training process uses the Backpropagation algorithm, which continuously adjusts the weights at each neuron's synapses. This approach enables the MLP to optimize its performance and learn complex data patterns effectively.

**LASSO:** Since its introduction by Tibshirani (1996), the Least Absolute Shrinkage and Selection Operator (LASSO) has become increasingly popular for its exceptional blend of feature selection and ridge regression benefits, primarily due to the development of effective algorithms. The idea is to constrain the sum of absolute regression coefficients, thereby shrinking specific coefficients and potentially eliminating insignificant variables, effectively achieving variable selection objectives. LASSO modeling can be outlined in the following way:

When presented with a dependent variable, *y*, and a set of independent variables, $x_1, x_2, \cdots, x_n$, the Ordinary Least Squares (OLS) approximation for the variable under study is:

$$\hat{y} = \beta_0 + \beta_1 x_1 + \beta_2 x_2 + \cdots + \beta_n x_n \quad (1)$$

LASSO extends OLS by incorporating a penalty term into the residual sum of squares (RSS). The penalty term $\alpha$ can be expressed as the product of the non-intercept beta coefficients' absolute values and a parameter λ, which regulates the speed of the penalty. For instance, when λ is less than 1, it decelerates the penalty, whereas values above 1 accelerate it.

$$Min \sum(RSS + \lambda \sum_{i=1}^{n} |\beta_i|) \quad (2)$$

Another way to rephrase LASSO is to minimize the RSS while constraining the sum of the absolute values of the non-intercept beta coefficients. It must not exceed a certain threshold. The beta coefficients decrease as s approaches 0, with less influential coefficients shrinking to zero before the more influential ones. Consequently, many beta coefficients that lack strong associations with the outcome are reduced to zero, removing their corresponding variables from the modeling. Thus, LASSO serves as a powerful variable selection method. Therefore, the following is a definition of the LASSO function.

$$Min \sum (y - \hat{y})^2 \tag{3}$$

$$\text{Subject to } \sum |\beta_i| \leq s \quad \text{where } i = 1 \cdots, n \tag{4}$$

As we decrease the value of $s$, certain $\beta_i$ values are forced to zero, removing the corresponding variables from the model.

**ECOC:** In ECOC, each class label is displayed through a unique dual code, and a set of dual sorters is trained, with each classifier distinguishing between one class and the rest (one-vs-all strategy) or between pairs of classes (one-vs-one strategy). After training the binary classifiers, the sorters' outputs are integrated to produce the final estimate. This combination can be achieved using techniques such as "voting" or "weighted voting," where each binary classifier's output contributes to the final decision. One of the primary benefits of ECOC is its ability to correct errors. Even if some binary classifiers make incorrect estimations, the final decision can still be made if the errors are "corrected" by the outputs of other classifiers. The efficiency of the ECOC encoding scheme relies on the Hamming distance between the binary codes assigned to different classes. A higher Hamming distance ensures greater "separation" between classes, making it easier for the binary classifiers to distinguish between them. Using Matrix $A$, we can define the Hamming distance $d_H(x, y)$ as:

$$d_H(x, y) = \sum_{i=0}^{k-1} \sum_{2j=0, j \neq i}^{k-1} a_{ij} \tag{5}$$

The total of each off-diagonal component of $A$ indicates the positions where $\boldsymbol{x}$ and $\boldsymbol{y}$ differ. The computational complexity of ECOC depends on factors such as the number of classes, the number of binary classifiers, and the complexity of the base sorters. Theoretical analysis can provide bounds on the computational complexity of ECOC algorithms (Danoyan, 2017).

Both ordered probit and ordered logit models share a common theoretical foundation based on an unobserved continuous latent variable (Williams & Quiroz, 2019; Stanford University, n.d.)

Observed ordinal variable: $Y_i \in \{1,2,\ldots,M\}$ with $M$ ordered categories

Unobserved latent variable: $Y_i^*$ (continuous, unbounded)

Threshold parameters: $\kappa_1 < \kappa_2 < \cdots < \kappa_{M-1}$

Predictor variables: $\mathrm{X}_i = (X_{i1}, X_{i2}, \ldots, X_{iK})^T$

Regression coefficients: $\beta = (\beta_1, \beta_2, \ldots, \beta_K)^T$

The latent variable is specified as a linear function of predictors plus a random disturbance term:

$Y_i^* = \mathrm{X}_i^T \beta + \epsilon_i = \sum_{k=1}^{K} X_{ik}\beta_k + \epsilon_i$

where $\epsilon_i$ is the error term whose distribution distinguishes ordered probit from ordered logit.

Note: The model typically does NOT include an intercept term in $\mathrm{X}_i^T \beta$, as the threshold parameters $\kappa_j$ serve this role (Aitchison & Silvey, 1957).

The observed ordinal outcome $Y_i$ is determined by which threshold region the latent variable $Y_i^*$ falls into: $Y_i = j \iff \kappa_{j-1} < Y_i^* \leq \kappa_j$

where $j = 1,2,\ldots,M$ and by convention:

- $\kappa_0 = -\infty$
- $\kappa_M = +\infty$

Explicit threshold conditions:

- $Y_i = 1$ if $Y_i^* \leq \kappa_1$
- $Y_i = 2$ if $\kappa_1 < Y_i^* \leq \kappa_2$
- $Y_i = 3$ if $\kappa_2 < Y_i^* \leq \kappa_3$
- $\vdots$

- $Y_i = M$ if $Y_i^* > \kappa_{M-1}$

Define the systematic component (linear predictor):

$$Z_i = X_i^T\beta = \sum_{k=1}^{K} X_{ik}\beta_k$$

This represents the expected value of the latent variable: $E(Y_i^*) = Z_i$

**Ordered Logit Model (Proportional Odds Model)**

In the ordered logit model, the error term follows a standard logistic distribution:

$$\epsilon_i \sim \text{Logistic}(0, \sigma^2 = \pi^2/3 \approx 3.29)$$

The standard logistic distribution has:

- Mean: $\mu = 0$
- Variance: $\sigma^2 = \pi^2/3 \approx 3.29$
- Cumulative distribution function (CDF): $\Lambda(z) = \frac{1}{1+e^{-z}} = \frac{e^z}{1+e^z}$

For ordered logit, the probability that an observation $i$ falls into the category $j$ is:

$$P(Y_i = j \mid X_i) = \Lambda(\kappa_j - Z_i) - \Lambda(\kappa_{j-1} - Z_i)$$

where $\Lambda(\cdot)$ is the logistic CDF (also called the inverse logit or expit function):

$$\Lambda(z) = \frac{1}{1+\exp(-z)} = \frac{\exp(z)}{1+\exp(z)}$$

Expanded form:

$$P(Y_i = j \mid X_i) = \frac{1}{1+\exp[-(\kappa_j - Z_i)]} - \frac{1}{1+\exp[-(\kappa_{j-1} - Z_i)]}$$

For the lowest category ($j = 1$):

$$P(Y_i = 1 \mid X_i) = \Lambda(\kappa_1 - Z_i) = \frac{1}{1+\exp(Z_i - \kappa_1)}$$

For middle categories ($j = 2,3, \dots, M-1$):

$$P(Y_i = j \mid X_i) = \Lambda(\kappa_j - Z_i) - \Lambda(\kappa_{j-1} - Z_i)$$

$$= \frac{1}{1+\exp(Z_i - \kappa_j)} - \frac{1}{1+\exp(Z_i - \kappa_{j-1})}$$

For the highest category ($j = M$):

$$P(Y_i = M \mid X_i) = 1 - \Lambda(\kappa_{M-1} - Z_i) = \frac{1}{1+\exp(\kappa_{M-1} - Z_i)}$$

When there are three ordered categories ($M = 3$), we estimate two thresholds ($\kappa_1, \kappa_2$):

$$P(Y_i = 1) = \frac{1}{1+\exp(Z_i - \kappa_1)}$$

$$P(Y_i = 2) = \frac{1}{1+\exp(Z_i - \kappa_2)} - \frac{1}{1+\exp(Z_i - \kappa_1)}$$

$$P(Y_i = 3) = 1 - \frac{1}{1+\exp(Z_i - \kappa_2)}$$

The ordered logit model is also called the proportional odds model because the odds of being in category $j$ or below versus above category $j$ are:

$$\frac{P(Y_i \le j)}{P(Y_i > j)} = \exp(\kappa_j - Z_i)$$

Taking logarithms:

$$\log\left[\frac{P(Y_i \le j)}{P(Y_i > j)}\right] = \kappa_j - Z_i$$

The effect of a one-unit increases in $X_{ik}$ on the log-odds is $-\beta_k$, which is constant across all thresholds (proportional odds assumption) (Menard, 2002).

**Ordered Probit Model**

In the ordered probit model, the error term follows a standard normal distribution:

$$\epsilon_i \sim N(0,1)$$

The standard normal distribution has:

- Mean: $\mu = 0$
- Variance: $\sigma^2 = 1$
- Cumulative distribution function (CDF): $\Phi(z) = \int_{-\infty}^{z} \frac{1}{\sqrt{2\pi}} e^{-t^2/2} dt$

For ordered probit, the probability that observation $i$ falls into category $j$ is:

$$P(Y_i = j \mid \mathbf{X}_i) = \Phi(\kappa_j - Z_i) - \Phi(\kappa_{j-1} - Z_i)$$

where $\Phi(\cdot)$ is the standard normal CDF.

For the lowest category ($j = 1$):

$$P(Y_i = 1 \mid \mathrm{X}_i) = \Phi(\kappa_1 - Z_i)$$

For middle categories ($j = 2,3, \dots, M-1$):

$$P(Y_i = j \mid \mathrm{X}_i) = \Phi(\kappa_j - Z_i) - \Phi(\kappa_{j-1} - Z_i)$$

For the highest category ($j = M$):

$$P(Y_i = M \mid \mathrm{X}_i) = 1 - \Phi(\kappa_{M-1} - Z_i)$$

When there are three ordered categories ($M = 3$), we estimate two thresholds ($\kappa_1, \kappa_2$):

$$P(Y_i = 1) = \Phi(\kappa_1 - Z_i)$$

$$P(Y_i = 2) = \Phi(\kappa_2 - Z_i) - \Phi(\kappa_1 - Z_i)$$

$$P(Y_i = 3) = 1 - \Phi(\kappa_2 - Z_i)$$

**Implementation as Scikit-Learn Classifier**

For integration into scikit-learn pipelines, implement:

`fit(X, y) method:`

1. Check input validity
2. Initialize parameters
3. Maximize log-likelihood via numerical optimization
4. Store $\hat{\beta}$ and $\{\hat{\kappa}_j\}$
5. Store unique class labels as `self.classes_`

`predict_proba(X) method:`

1. Compute $Z_i = \mathrm{X}_i^T \hat{\beta}$ for each observation
2. Evaluate $F(\hat{\kappa}_j - Z_i)$ for all thresholds
3. Return probability matrix of shape $(N, M)$

`predict(X) method:`

1. Call `predict_proba(X)` to get probability matrix
2. Return `self.classes_[np.argmax(probs, axis=1)]`

**Example Application: Credit Rating Classification**

Consider a bankruptcy prediction context with ordinal credit ratings: Healthy (1), Watch (2), Distressed (3).

Predictor variables:

- $X_1$: Debt-to-equity ratio
- $X_2$: Current ratio

- $X_3$: Return on assets

Ordered logit specification:

$$Y_i^* = \beta_1 X_{i1} + \beta_2 X_{i2} + \beta_3 X_{i3} + \epsilon_i, \epsilon_i \sim \text{Logistic}(0, \pi^2/3)$$

- $Y_i = 1$ (Healthy) if $Y_i^* \leq \kappa_1$
- $Y_i = 2$ (Watch) if $\kappa_1 < Y_i^* \leq \kappa_2$
- $Y_i = 3$ (Distressed) if $Y_i^* > \kappa_2$

Interpretation:

- Negative $\hat{\beta}_1$: Higher debt-to-equity increases probability of worse rating
- Positive $\hat{\beta}_2$: Higher current ratio increases probability of better rating
- Positive $\hat{\beta}_3$: Higher ROA increases probability of better rating

Prediction:
For a new firm with $(X_1, X_2, X_3) = (2.5, 1.8, 0.05)$:

$$\hat{Z} = \hat{\beta}_1(2.5) + \hat{\beta}_2(1.8) + \hat{\beta}_3(0.05)$$

$$\hat{P}(Y = 1) = \frac{1}{1 + \exp\left(\hat{Z} - \hat{\kappa}_1\right)}$$

$$\hat{P}(Y = 2) = \frac{1}{1 + \exp\left(\hat{Z} - \hat{\kappa}_2\right)} - \frac{1}{1 + \exp\left(\hat{Z} - \hat{\kappa}_1\right)}$$

$$\hat{P}(Y = 3) = 1 - \frac{1}{1 + \exp\left(\hat{Z} - \hat{\kappa}_2\right)}$$

Assign to the category with the highest probability.

## C. Performance Metrics and Feature Importance of Datasets

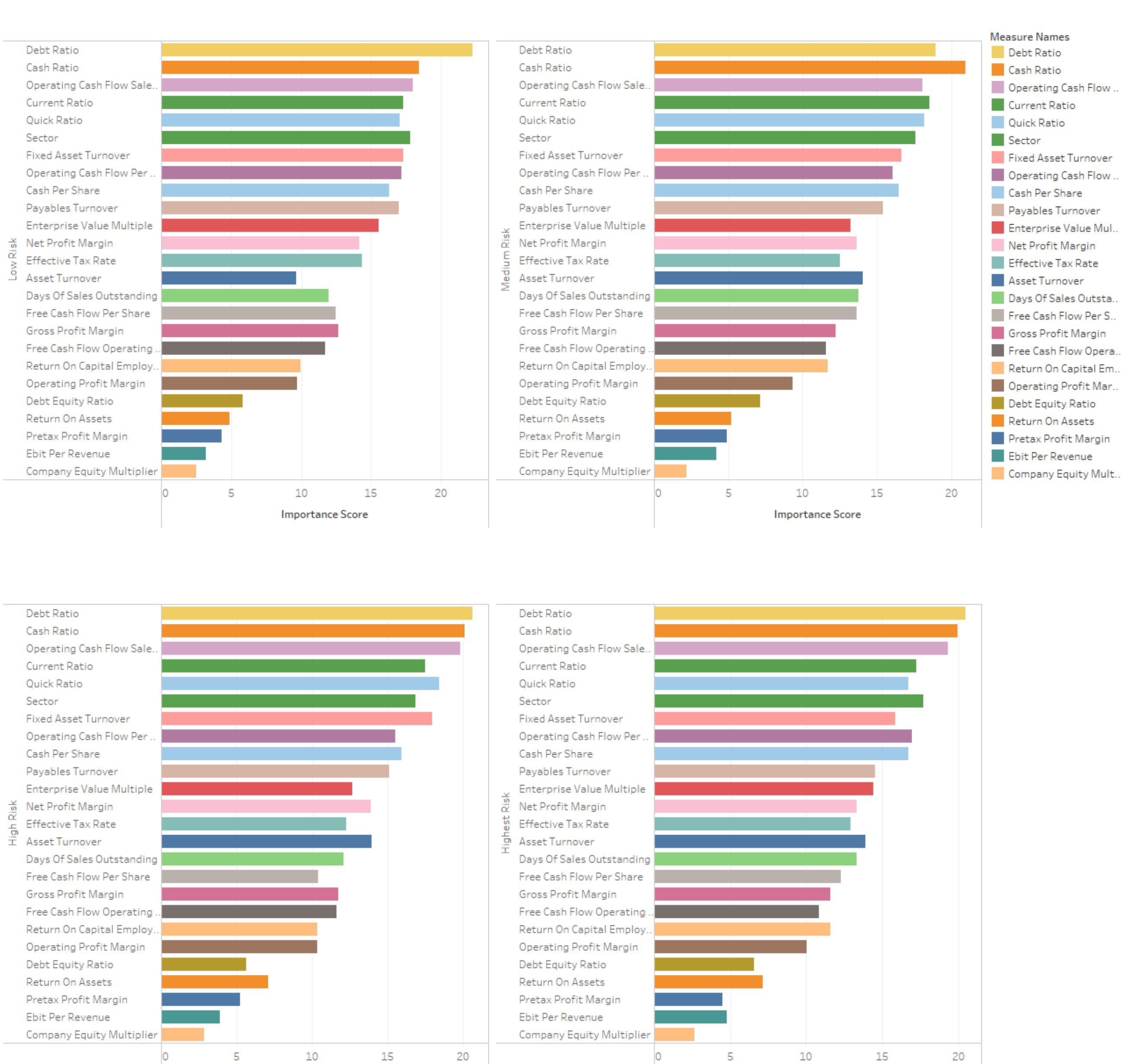

Figure 1a. Feature importance score for Dataset 1 at various risk levels.

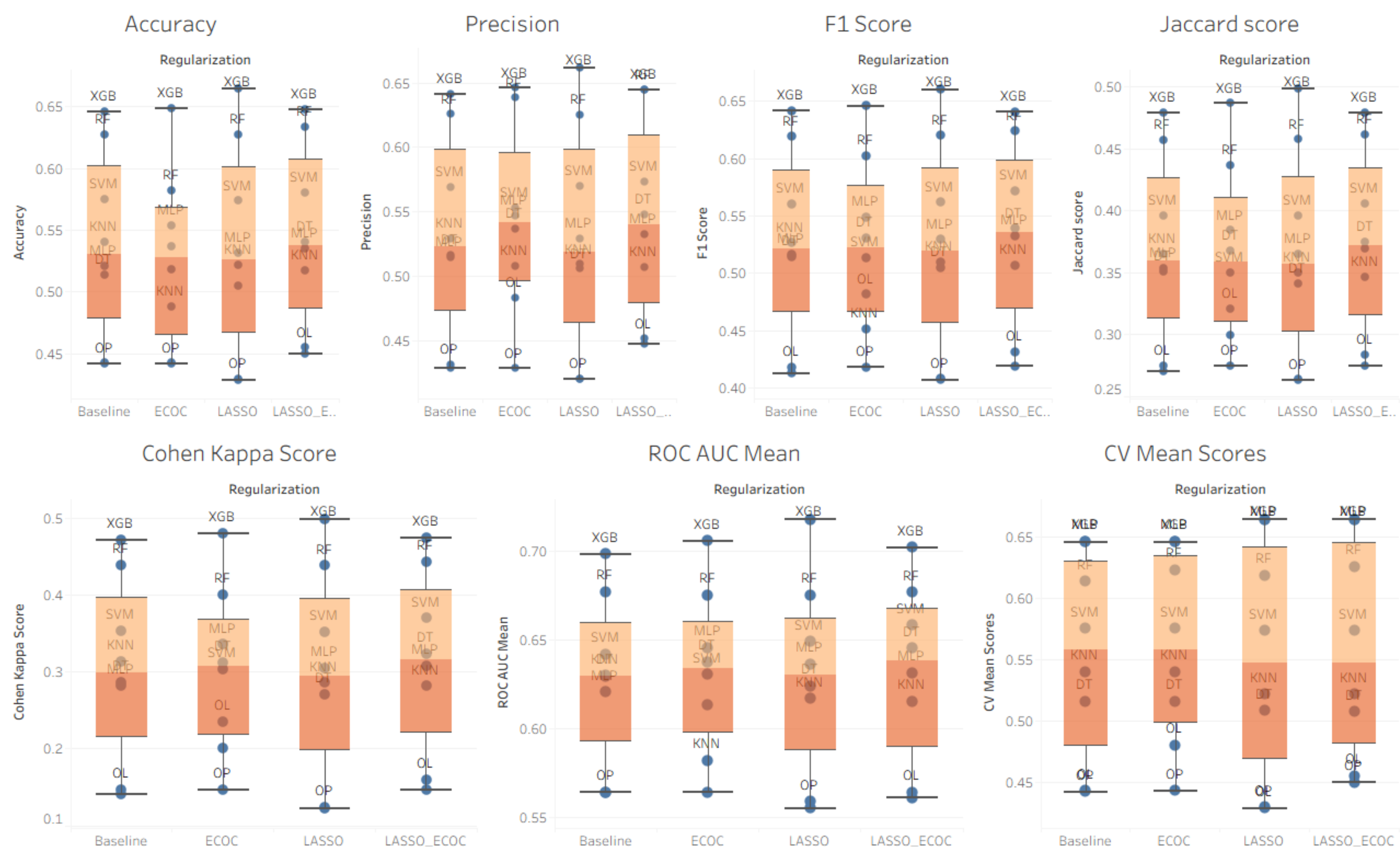

Figure 2a. Metric comparison on different regularization models for Dataset 1.